\documentclass[letterpaper, 10 pt, conference]{ieeeconf}  

\IEEEoverridecommandlockouts                              

\usepackage{graphicx} 
\usepackage{times} 
\usepackage{amsmath} 
\usepackage{amssymb}  
\usepackage{url}
\usepackage{cite}
\usepackage{enumerate}
\usepackage{bm}
\usepackage{cases}
\usepackage{mathtools}
\usepackage[caption=false, font=footnotesize, labelfont=sf,textfont=sf]{subfig}
\usepackage{algorithmicx}
\usepackage{algorithm}
\usepackage{algpseudocode}
\usepackage{booktabs, ragged2e}
\usepackage{tabularx}
\usepackage{graphicx}
\usepackage{multirow}
\usepackage[para]{threeparttable}
\usepackage{caption}
\usepackage[table,xcdraw]{xcolor}

\makeatletter
\let\NAT@parse\undefined
\makeatother
\usepackage[hidelinks]{hyperref} 

\newcolumntype{C}{>{\Centering\arraybackslash}X}
\newcolumntype{L}{>{\raggedright\arraybackslash}X}
\newcolumntype{R}{>{\raggedleft\arraybackslash}X}

\title{\LARGE \bf RiskMap: A Unified Driving Context Representation for\\ Autonomous Motion Planning in Urban Driving Environment}

\author{
Ren Xin, Sheng Wang, Yingbing Chen, Jie Cheng, Ming Liu, and Jun Ma
\thanks{This work was supported by the Guangzhou-HKUST(GZ) Joint Funding Scheme under Grant 2024A03J0618. \textit{(Corresponding author: Jun Ma.)}}
\thanks{Ren Xin, Sheng Wang, Yingbing Chen, and Jun Ma are with the Division of Emerging Interdisciplinary Areas, The Hong Kong University of Science and Technology, Hong Kong SAR, China, and also with the Robotics and Autonomous Systems Thrust, The Hong Kong University of Science and Technology~(Guangzhou), Guangzhou 511453, China. (e-mail: \{rxin, swangei, ychengz\}@connect.ust.hk; jun.ma@ust.hk)}
\thanks{Jie Cheng is with the Department of Electronic and Computer Engineering, The Hong Kong University of Science and Technology, Hong Kong SAR, China. (e-mail: jchengai@connect.ust.hk)}
\thanks{Ming Liu is with the Robotics and Autonomous Systems Thrust, The Hong Kong University of Science and Technology~(Guangzhou), Guangzhou 511453, China. (e-mail: eelium@hkust-gz.edu.cn)}}

\begin{document}
\maketitle
\thispagestyle{empty}
\pagestyle{empty}
\begin{abstract}
        Motion planning is a complicated task that requires the combination of perception, map information integration and prediction, particularly when driving in heavy traffic.
        Developing an extensible and efficient representation that visualizes sensor noise and provides basis to real-time planning tasks is desirable.   
        We aim to develop an interpretable map representation, which offers prior of driving cost in planning tasks. 
        In this way, we can simplify the planning process for dealing with complex driving scenarios and visualize sensor noise.
        Specifically, we propose a unified context representation empowered by deep neural networks.
        The unified representation is a differentiable risk field, which is an analytical representation of statistical cognition regarding traffic participants for downstream planning tasks. 
        This representation method is nominated as RiskMap.
        A sampling-based planner is adopted to train and compare RiskMap generation methods. 
        In this paper, the RiskMap generation tools and model structures are explored, the results illustrate that our method can improve driving safety and smoothness, and the limitation of our method is also discussed.

\end{abstract}


\section{Introduction}

Deep Neural Network~(DNN) is widely used to approximate high-dimensional feature functions that are hard to mathematically characterize. 
It has been proven to be able to approximate any Lebesgue integrable function with sufficient parameters and a suitable structure~\cite{NIPS2017_LU}. However, millions of parameters make DNN difficult to analyze and interpret. 
At the same time, traditional analytical planning methods, including optimization and sampling, are only reliable in limited scenarios. 
This paper presents an effective and interpretable method for combining the advantage of traditional and deep learning-based approaches.

\subsection{Motivation}

Motion planning under constraints resulting from the traffic environment and the predicted movements of other vehicles is a typical strategy for autonomous vehicles~(AV). 
For prediction, a network-based probabilistic model seems more competent as the maneuver of the agent in traffic is full of uncertainties. 
It is possible that at some point other traffic participants will do something outside the rules.
However, the planning module, as an actor, should not make maneuvers beyond the rules.
But an end-to-end neural network planner is hard to treat curb or traffic rules as a hard constraint. 
It is commonly observed to lead to rule violating results.
Thus, the planning process may be considered a deterministic decision process regulated by hard constraints, rather than a probabilistic model.

In a prediction-planning decoupling framework, ensuring the rich and efficient expression of prediction results~\cite{velocityfield,occflow}, as well as the sufficient interpretability and flexibility of the planning module~\cite{mp3,CIOC,diff_MPC} are important. 
Therefore, methods for mapping circumstance information to risk space and best-cost trajectory selection mechanisms are explored thoroughly. 
 
\begin{figure}
  \centering
  \includegraphics[width=3.3in]{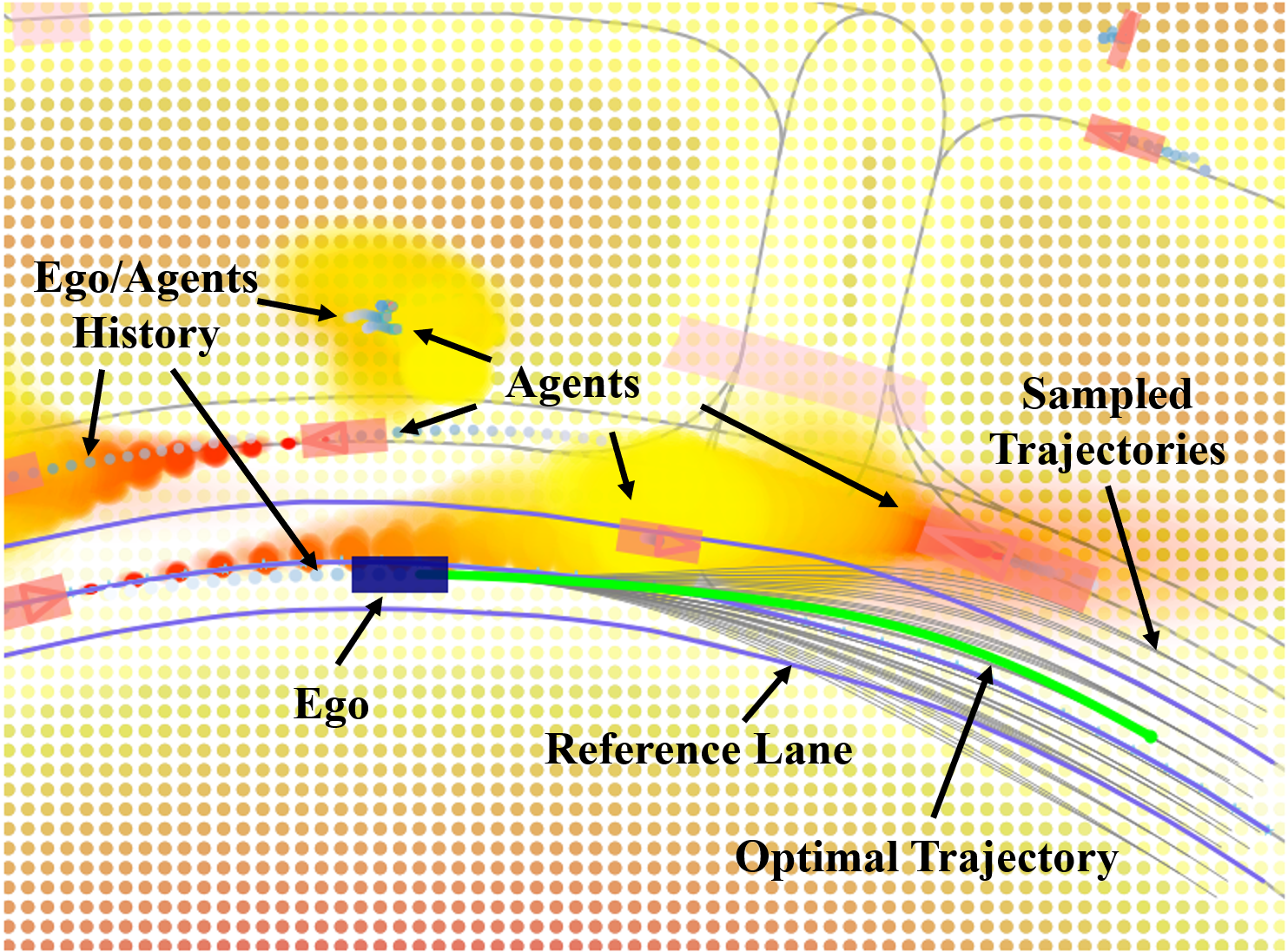}
  \caption{
    Planning in urban scenarios is a challenging task because it needs to consider the perception, prediction, and control uncertainties together with complex road environments. 
    Our method encodes and represents all perception information in a risk map, which is a middleware of sensing and planning.
    The purple rectangle represents the ego vehicle, the red blocks are other agents. 
    The transparency of \textbf{orange-yellow} fields represent the occupancy probability of other traffic participants. 
    The \textbf{white-yellow-orange} points represent the risk value of sampled positions. 
    Sample points with deeper shades of red signifying more severe risks.
  }
  \label{fig:cover_img}
\end{figure}

\subsection{Challenges}




\textbf{The Representation and Utilization of Prediction Results}. 
Modern HD map, cognition, and prediction methods offer plentiful driving context information, including reference lanes, traffic lights, occupancy maps, and multi-modal future trajectories with probability.
Interactive planning methods with multi-modal prediction are inspected in~\cite{risk_boris, MAGIC}. 
Previous works~\cite{mp3,occflow} draw on the concept of flow commonly used in motion perception to predict the position of other traffic participants, while no explicit utilization is developed.

\textbf{Regularization of Cost Function}.
To address the aforementioned problems, the deep learning-based cost function is introduced. 
However, directly minimizing the cost of expert trajectories will result in letting all the sampled states at a small value, rather than contributing to indicate which trajectory is optimum. 
Therefore, loss terms regularizing the cost value is in need for avoiding the model finding the shortcut.

\subsection{Contributions}
To address the above issues and increase flexibility and interpretability in planning, we propose the RiskMap framework. To the best knowledge of us, this method is the first to generate interpretable unified driving context representation in risk space by imitation learning.
Our contributions are mainly threefold:

(1) We propose a novel generator of differentiable signed risk field, which is an analytical representation of probabilistic prediction distribution for downstream planning tasks.

(2) We develop a real-time planner that generates smooth, human-like trajectories with RiskMap output.

(3) We integrate RiskMap and the planner into a unified framework, which demonstrates the efficiency of map-generating modules and the effectiveness of modeling tools.

\section{Related Work}






\subsection{Driving Motion Prediction}
Traditional rule-based behavior prediction methods generate multiple possible trajectories based on constraints from road maps. 
Thanks to the great development of deep learning, more recently, many learning-based approaches are proposed~\cite{mtp,CoverNet}, which take the rasterized map information and vehicle history trajectory as input and output different hypotheses meanwhile assigning the possibilities to each of them.
%
Some other methods~\cite{vectornet,PGP} choose to encode contextual information in vector format which is more powerful since its more efficient context representation ability. 
They use the attention mechanism or graph convolution method to aggregate information and learn the latent interaction model so that the predictor can pay more attention to the information most related.
%
In addition, Jie \textit{et al.}~\cite{cj2022mpnp} emphasizes reference lane in context presentation to enhance the success rate and diversity of planning. 

\subsection{Representation of Map Information}
The map is of vital importance for mobile robots~\cite{jiao2022fusionportable} as it is an integration cognition result.
The concept of occupancy grid map~\cite{OCC_Map} and configuration space simplifies the problem of mobile robots in the past decades. 
The utilization of the signed distance field algorithm in 2D static maps efficiently offers gradient in trajectory optimization tasks. 
Rasterized probabilistic occupancy map~\cite{HOME,GOHOME} is adopted to improve the rationality of the distribution of the target position. 
HD maps~\cite{hdmap} offering a variety of semantic information is essential for modern urban autonomous driving. 
The occupancy flow is developed to predict the motion potential in a driving scenario~\cite{mp3,occflow} but, to our knowledge, no analytical planning algorithm based on it is demonstrated. 

On the other hand, the risk map is not a brand-new concept. Risk is defined as the collision probability in the future~\cite{risk_servey}. 
Nishimura \textit{et al.}~\cite{risk_boris} introduces prediction entropy as an optimization item of Model Predictive Control~(MPC). 
Huang \textit{et al.}~\cite{SafePlan} obtain obstacle uncertainty from traditional perception models and discusses combination methods of risk value of obstacles.

\subsection{Inverse Optimal Control and Imitation Learning}
There are several branches of learning from demonstration, including Behavior Cloning~(BC), Inverse Optimal Control~(IOC), and Inverse Reinforcement Learning~(IRL). 
IOC is to construct a cost function under a relatively stable system, while IRL is to recover a cost function based on a DNN in relatively complex systems~\cite{review_ioc}. 
Abel \textit{et al.}~\cite{yaman2013survey} first mathematically studies the inverse mechanical problems for finding the curve of an unknown path.
Discrete-time IOC scheme for trajectory tracking of non-linear systems is developed by~\cite{discreteIOC}.
Levine \textit{et al.}~\cite{CIOC,levine_nonlinear} represents the reward function as a Gaussian Process~(GP) that maps measurements from feature values to rewards, which inspired this work to use the concept of risk to uniform measurements.
Previous works~\cite{NMP,mp3,lookout} combine end-to-end learning with a classical sampling-based planner, either by estimating a cost volume or predicting diverse future trajectories of other agents. 
However, their results are not reproducible due to the use of a private dataset.

\section{Methodology}
\begin{figure*}[ht]
  \centering
  \includegraphics[width=\linewidth]{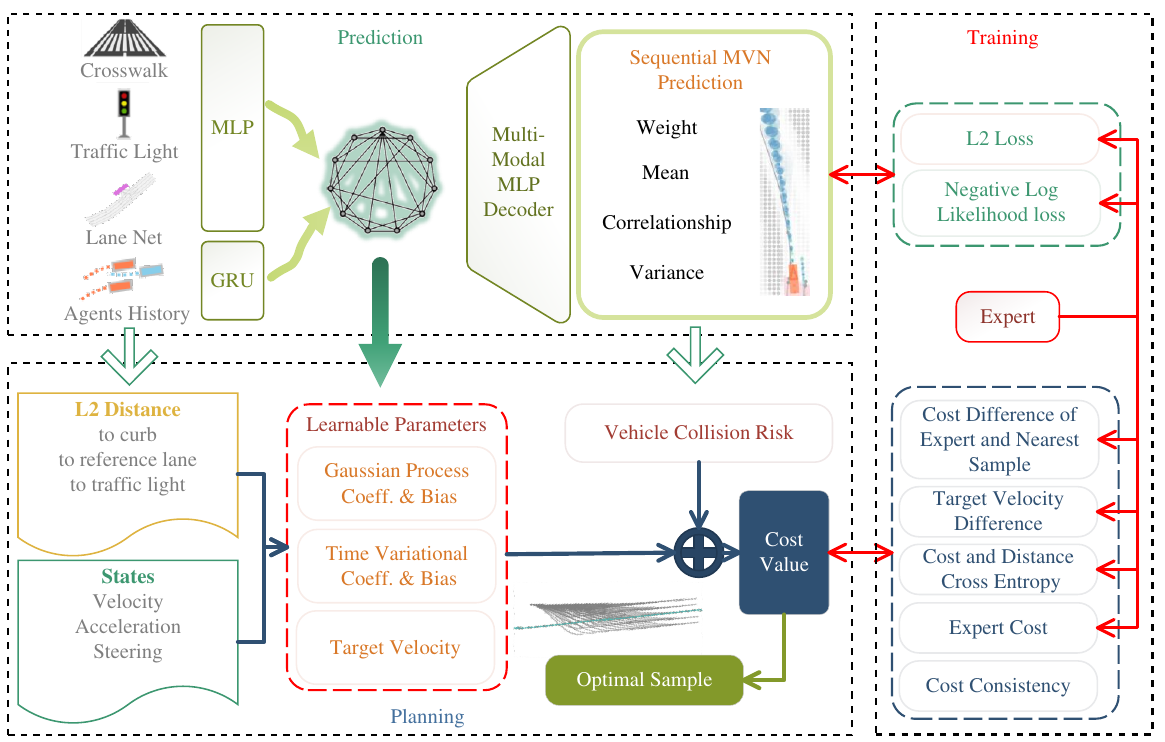}
  \caption{
    Illustration of the proposed framework of map generation and planning module with their training process. 
    The framework is roughly divided into two modules: prediction and planning. 
    \textbf{Prediction:} Reference lines, map elements, ego \& agents motion, and geometry are encoded with MLP/GRU modules. 
    Global attention is used to model the interaction between objects. 
    The multi-modal MLP decoder generates a series of multivariate normal distribution parameters to represent the vehicle future positions.
    \textbf{Planning:} A sampling-based planner is adopted, and all the sampled spatial-temporal points are with distance information to representative represent driving contexts.
    The global encoded latent variable in the prediction module generates parameters mapping distance to risk value space.
    Risk values are concatenated with collision risk. 
    The trajectory with the minimum risk is selected as the planning result.
    \textbf{Training:} A two-stage supervised learning process is adopted. The predictor and planner are trained with listed metrics, respectively.
  }
  \label{fig:overview}
\end{figure*}

\subsection{Problem Formulation}
This task is to generate a set of future trajectory points of the ego vehicle $\mathbf{P} = \left[p^1, \dots, p^T \right]$ in a short time period $T$ subject to history tracking data of other agents $h_a$ and ego vehicle $h_e$, map information of current time $\bm{\mathcal{M}}$ include reference lane $m_{ref}$, traffic light status $m_{tl}$, and static obstacles $m_{ob}$. 
The driving context is denoted as $\mathcal{S}  = (\bm{h_a},\bm{h_e},\bm{\mathcal{M}})$ and our model aims to find a mapping relationship $\mathbf{P} = \mathcal{F} (\mathcal{S})$.


\subsection{Driving Context Encoding} \label{sec:driving-context}
Similar to Vectornet~\cite{vectornet}, a graph neural network is implemented to encode vectorized context information for the prediction module. Both $T$ time step ego-vehicle history $h_e \in \mathbb{R}^{T\times 5}$ and $N_a$ surrounding vehicles $h_a \in \mathbb{R}^{N_a \times T \times 6}$ features are concatenated together and embedded by Gated Recurrent Unit~(GRU) to encode the time dimension information. 
Then the features extracted in 192-dimensional space are processed by a Multi-Layer Perception~(MLP). Map elements like piecewise lanes and crosswalk polyline are initially directly encoded by the MLP. 
All 192-dimensional encoded features are concatenated in the first dimension and aggregated by a Global multi-head ATtention network~(GAT) to get the vehicle-to-vehicle and vehicle-to-context relationships.


\subsection{Prediction Results Representation}
\begin{figure}
  \centering
  \includegraphics[width=\linewidth]{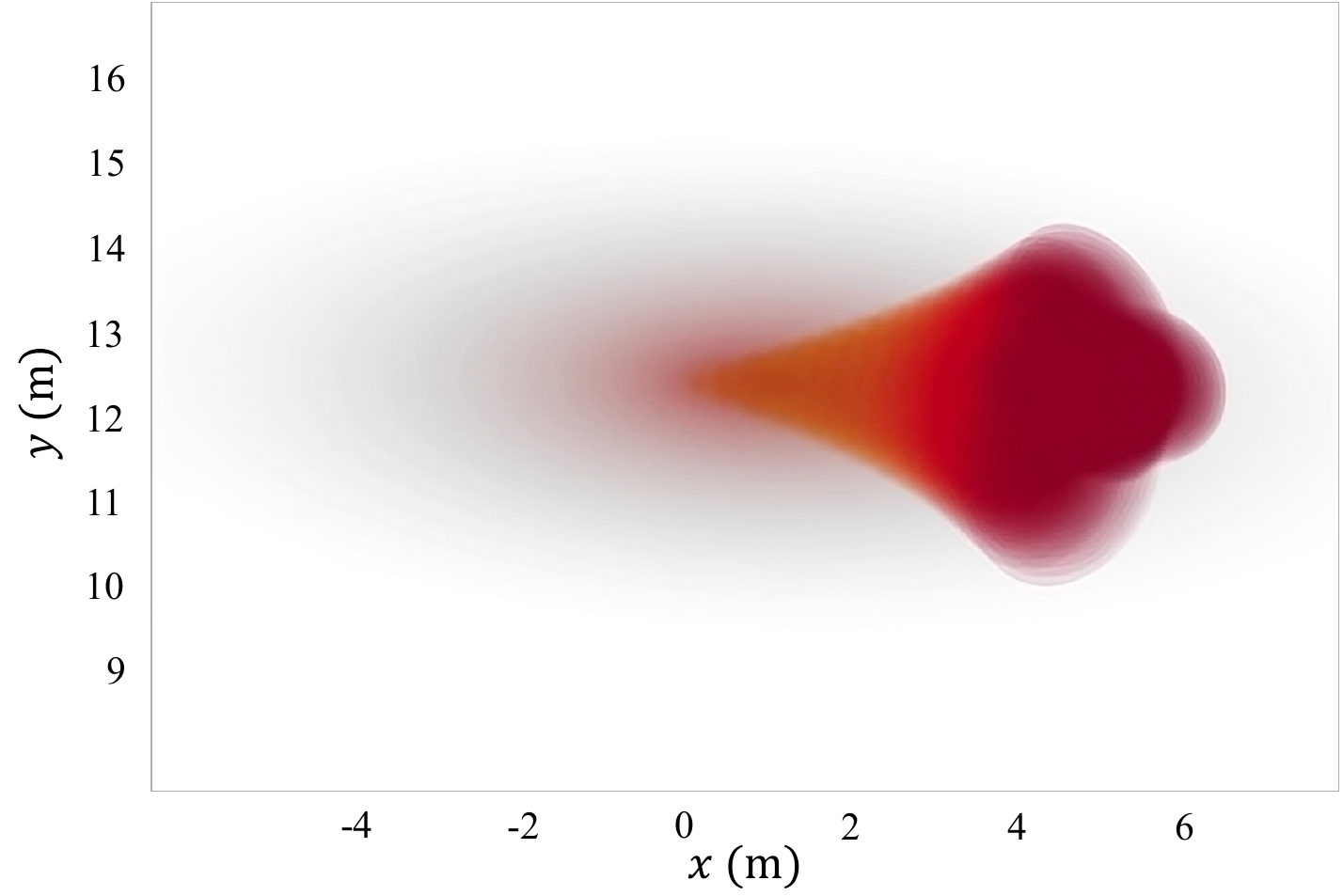} 

  \caption{Demonstration of SeqMVN, deeper color represents later time and the transparency represents the occupancy probability. The image shows that the multi-modality of SeqMVN helps to fit the complex shape of the probabilistic occupancy map. And the Gaussian distribution at different time points in each modal is correlated.
  }
  \label{fig:gmm}

\end{figure}

The presentation method is modeled as Sequential Multi-Variational Normal~(SeqMVN) distribution as Fig.~\ref{fig:gmm}. 
A tuple $\mathcal{T}_{mt} = (\mu_x, \mu_y, \sigma_x, \sigma_y, \rho)$ for each trajectory point is generated to present the predicted position center and uncertainty of a vehicle regarding (\ref{eq: mvn}). 
To ensure the prediction is temporal conditioned on each of their multi-modal results, the overall output is $\mathcal{T} \in \mathbb{R}^{N_a\times N_{modal} \times T \times dim(p_{mt})} $ and selection probability $Cls \in \mathbb{R}^{N_a \times N_{modal} }$ which can also be considered as weight of Gaussian Mixture Model~(GMM) components at a time step for one modal.
The likelihood of demo trajectory point $\mathcal{T}_{demo}^t$ is 
\begin{equation}
  \label{eq: mvn}
  f(\mathcal{T}_{demo}^t)= Cls_m^t\frac{\exp\left(
  -\frac 1 2 
  ({\mathcal{T}_{demo}^t}-{\boldsymbol\mu})^\mathrm{T}
  {\boldsymbol\Sigma}^{-1}
  ({\mathcal{T}_{demo}^t}-{\boldsymbol\mu})
  \right)}
  {\sqrt{(2\pi)^k |\boldsymbol\Sigma|}},
\end{equation}
where 
\begin{equation}
  \boldsymbol\mu = \begin{pmatrix} \mu_x \\ \mu_y \end{pmatrix}, \quad
  \boldsymbol\Sigma = \begin{pmatrix} \sigma_x^2 & \rho \sigma_x \sigma_y \\
                           \rho \sigma_x \sigma_y  & \sigma_y^2 \end{pmatrix},
\end{equation}
which will avoid the backward tracking problem of different agents~\cite{occflow}.
To make sure the distribution is non-degenerate, the output of the MLP-based decoder at the $\sigma$ and $\rho$ positions are regularized by exponential and hyperbolic tangents, respectively.

Sampling is essential for training the prediction model with uncertainty. 
Reparameterization is utilized to get the loss for each agent, similar to~\cite{mtp} to cover the gradient fault problem of sampling in backward propagation.
Negative Log Likelihood~(NLL) loss is adopted.
The likelihood of ground truth points of an agent is $\mathcal{P}^a_{mt}\in\mathbb{R}^{N_a \times M \times T} $ is calculated by (\ref{eq: mvn}). 
Vanish of each letter represents summation up along that dimension, sum along time horizon is represented as $\mathcal{P}^a_{m}\in \mathbb{R}^{N_a \times M}$. 
The optimization target of the prediction stage of each agent is defined as
\begin{equation}
  \begin{aligned}    
    \mathcal{L}_{pre}^a &= \\
    \sum_{m = 1}^M &\left( -\log(\mathcal{P}^{a}_{m}) 
    -\log\left(
    Cls_m^a
    \left[
    \min(
    \lVert {
    \mathcal{T}_{pre}^{a} 
    -\mathcal{T}_{demo}^{a} 
    }\rVert 
    _{2})
    \right]
    \right)
    \right),
  \end{aligned}
\end{equation}
where the first item means minimum NLL between expert trajectory and predicted trajectory, and the second item means maximizing the probability of selecting the closest prediction modal.


\subsection{Measurement of Driving Context}
The boundary of the ego vehicle is presented by a multi-circle model according to their lengths, which means all distances are calculated with several circles representing the outline of traffic participants. 
Three Euler distances are measured, the distance to the closest reference lane $d_{ref}$, the distance to the traffic light $d_{tl}$, including the distance to static traffic obstacles $d_{sdf}$. The raw measurements are denoted as $\mathcal{D}  = [d_{ref},d_{sdf},d_{tl}]^\top$. The elements are mapped to the value space of the signed risk map by
\begin{equation}
  \label{eq:maprisk}
  \mathcal{R}_\mathcal{D} = \beta \cdot \exp\left( \lambda \mathcal{D} \right),
\end{equation}
where $\beta$ and $\lambda$ are generated by an MLP regarding each encoded context feature.
The probability of collision with dynamic vehicles of each sample at time step t is 
\begin{equation}
  \label{eq:colli}
  \begin{split}
    \mathcal{R}_{col} = \mathcal{P}_{t}. 
  \end{split}
\end{equation}
The concatenated four dimensional vector $\mathcal{R} = \{\mathcal{R}_\mathcal{D}, \mathcal{R}_{col}\}$ is nominated as risk raw measurements on each sampled trajectory point.

\subsection{Design of Cost Function}
Weight vectors are generated by MLPs regarding each aggregated context feature. 
Symbol $w_{smooth}$ represents that preference of driving smoothness $C_{smooth} = (a,s) \cdot w_{smooth}$. 
Second-order smoothness parameters, including acceleration $a$ and steering angle $s$, represents yaw rate differed by ground truth and sampled trajectory. 
The sampled trajectories are generated by a lattice planner~\cite{optimalTraj,lattice}.
Difference between expected target velocity $\bar{v}$ and current velocity $ v$
$
  d_v = \bar{v} - v  
$
is also taken into consideration, where $\bar{v}$ is generated by a swallow decoder.
The measurement feature vector is concatenated together. The cost of each trajectory is 
\begin{equation}
  \mathcal{C}_{\mathcal{S}} = cat(\mathcal{R},C_{smooth},d_v \cdot w_d),
\end{equation}
where $w_d$ is a coefficient of velocity difference learned by the model.

\subsection{Loss Definition During Training Process}
Excluding minimizing the driving cost, expected velocity $\bar{v}$ is individually regressed by 
\begin{equation}
  \label{eq:idr}
  \mathcal{L}_v = \sum_{t = 0}^{T} (\bar{v}^t - v^{t}_{\mathcal{T}_{demo}})^2,
\end{equation}
which is designed for avoiding freezing robot status.
The selection item calculates Cross Entropy~(CE) of cost distribution and distance distribution of sampling trajectory set $\mathcal{T}_s $ by  
\begin{equation}
  \mathcal{L}_{sel} = CE\left(Softmin(\mathcal{C}_{\mathcal{T}_s}),Softmin(\left\lVert\mathcal{S}_s-\mathcal{T}_{demo}\right\rVert _2)\right).
\end{equation} 
Regression item finds the cost difference between selected trajectory and expert trajectory by
\begin{equation}
  \mathcal{L}_{l2} = \sum_{t = 0}^{T} \left( \mathcal{C}_{\mathcal{T}^{t}_{nearest}}-\mathcal{C}_{\mathcal{T}_{demo}^t} \right).
\end{equation}
However, all the above loss items will mislead the model to get similar cost values. So a loss item minimizing cost consistency is designed as 
\begin{equation}
  \mathcal{L}_{con} = \operatorname{Var}(\mathcal{C}_\mathcal{T})^{-1}.
\end{equation}
To this end, the overall loss is 
\begin{equation}
  \mathcal{L} = \mathcal{C}_{\mathcal{T}_{demo}} + \mathcal{L}_{sel} + \mathcal{L}_{l2} + \mathcal{L}_{con} + \mathcal{L}_v.
\end{equation}

\subsection{Training and Inference Details} \label{sec:aug}
A two-stage supervised learning process is conducted. We first train the prediction model from all the vehicles, like the general multiagent training process. Then the map parameter and cost function training process is conducted by imitating the behavior demo of the ego vehicle.
The raw context information is processed the same as in the first stage, and prediction results are input into the second stage.

\section{Experiment}
\begin{table}[]
  \centering
  \caption{Comparison with Imitation Learning}
  \renewcommand\tabcolsep{8.0pt}
  \label{tab:compare}
  \begin{tabular}{@{}l|c|ccc|c@{}}
  \toprule
  \multicolumn{1}{c|}{\multirow{2}{*}{Method}} & \multicolumn{1}{c|}{ADE} & \multicolumn{3}{c|}{Collision Rate~(\%)}      & Jerk~$(\textup{m/s}^3)$ \\
  \multicolumn{1}{c|}{}                        & (m)                   & 1.0\,s          & 2.0\,s          & 3.0\,s          & 0-3.0\,s                       \\ \midrule
  ChaufferNet                                  & 0.943                    & 0.33          & 0.56          & 0.83          & 1.917                        \\
  ChaufferNet-Vec                              & \textbf{0.475}           & 0.31          & \textbf{0.49} & \textbf{0.71} & 2.679                        \\
  RiskMap~(ours)                                & 1.084                    & \textbf{0.00} & 1.31          & 3.17          & \textbf{0.2107}              \\ \bottomrule
  \end{tabular}
  \end{table}
\begin{table}[]
  \centering
  \caption{Ablation Study of Model}
  \label{tab:ablation}
    \begin{tabular}{@{}cccc|c|cc|c@{}}
    \toprule
    Risk      & LN        & TV        & \begin{tabular}[c]{@{}c@{}}Samp. \\ Count\end{tabular} & \begin{tabular}[c]{@{}c@{}}ADE\\ (m)\end{tabular} & \begin{tabular}[c]{@{}c@{}}FDE\\ Lat.~(m)\end{tabular} & \begin{tabular}[c]{@{}c@{}}FDE\\ Lon.~(m)\end{tabular} & \begin{tabular}[c]{@{}c@{}}Jerk\\ $(\textup{m/s}^3)$\end{tabular} \\ \midrule
    \checkmark &           &           & 400                                                    & 1.16                                              & \textbf{1.48}                                         & 2.01                                                  & \textbf{0.2060}                                                                  \\
              & \checkmark & \checkmark & 400                                                    & 1.99                                              & 1.63                                                  & 3.78                                                  & 0.2301                                                                  \\
    \checkmark &           & \checkmark & 100                                                    & 1.17                                              & 1.85                                                  & 1.84                                                  & 0.2115                                                                  \\
    \rowcolor[HTML]{EFEFEF} 
    \checkmark & \textbf{} & \checkmark & 400                                                    & 1.08                                              & 1.88                                                  & 1.57                                                  & 0.2107                                                                   \\
    \checkmark &           & \checkmark & 900                                                    & \textbf{1.07}                                     & 1.90                                                  & \textbf{1.53}                                         & 0.2111                                                                  \\ \bottomrule
    \end{tabular}
\end{table}
\begin{table}[t]
\centering
\caption{Ablation Study of Loss Items}
\label{tab:AblationLoss}
\renewcommand\tabcolsep{10.0pt}
\begin{tabular}{@{}l|c|cc|c@{}}
  \toprule
  Removed Loss Item & \begin{tabular}[c]{@{}c@{}}ADE\\ (m)\end{tabular} & \begin{tabular}[c]{@{}c@{}}FDE\\ Lat.~(m)\end{tabular} & \begin{tabular}[c]{@{}c@{}}FDE\\ Lon.~(m)\end{tabular} & \begin{tabular}[c]{@{}c@{}}Jerk\\ $(\textup{m/s}^3)$\end{tabular} \\ \midrule
  Ind. Vel. Reg.    & 6.37                                              & \textbf{0.67}                                         & 13                                                    & 0.2476                                                                  \\
  L2 Cross Entropy  & 1.32                                              & 1.95                                                  & 2.1                                                   & 0.2219                                                                  \\
  Cost Consistancy  & 1.26                                              & 2.28                                                  & 1.72                                                  & 0.2148                                                                  \\
  Cost Regression   & 1.03                                              & 1.69                                                  & \textbf{1.61}                                         & 0.2106                                                                  \\
  Demo Cost         & \textbf{0.98}                                     & 1.29                                                  & 1.67                                                  & \textbf{0.2030}                                                          \\ \bottomrule
  \end{tabular}
\end{table}

\subsection{Implementation Details}


\textit{(1) Dataset:}
Both the two training stages of our model are carried out on the Lyft Motion Prediction Dataset~\cite{lyft_dataset}, which contains 134\,k, 25\,s time horizon, real-world driving samples, collected on a complex urban route in Palo Alto, California.
HD maps and annotations for traffic signals/participants are provided at 10~Hz.
A subset of about 3.6\,M frames is randomly selected for training and tested on a subset of the original validation set which contains 3600 frames.

\textit{(2) Training:}
We train the prediction model using the AdamW~\cite{adamw} optimizer using a batch size of 128 and a learning rate of 0.0001 for 18 epochs on a single RTX 3080 laptop GPU.
The RiskMap planning model is trained on 1/10 of the subset for one epoch on the same device.
No dataset augmentation is implemented in either of the training stages.
We train both the prediction model and RiskMap model to show trajectories for 3\,s in the future, given the last 1.5\,s of observation.

\textit{(3) Evaluation:}
We run our model and other baselines on the logged data and report several common metrics including Average Distance Error~(ADE), Final Distance Error~(FDE), Lat. and Lon. are short for lateral and longitudinal offset, Collision Rate, and Jerk. The average operation time of generating a planning result is 0.048\,s~(Min. 0.017, Max. 0.111). The operating time meets the real-time requirement. 


\subsection{Comparison with Baselines}
We compare the planning results against two representative baselines on the Lyft dataset.
\textbf{\textit{ChauffeurNet}}~\cite{Chauffeurnet}, where the input is rasterized images and perturbation is used to deal with co-variate shift.
\textbf{\textit{ChaufferNet-Vec}}, where the rasterized image input is replaced with the more advanced vectorized representation~\cite{vectornet}.
The comparison result are listed in Table~\ref{tab:compare}. The baselines are selected to show imitation learning performance. 
From Table~\ref{tab:compare} we know our method has the lowest jerk value, as it is strictly following dynamic rules and the sampling module offers jerk optimized trajectories.
The capability of dealing with collision and copying expert behavior is not surpassing learning-based methods no matter what kind of cost remapping method is adopted. 
The results illustrate that pure distance information may not sufficient for drivers decision. 
But it still achieves comparable performance with better interpretability.

\subsection{Ablation Study of Model Components}

Table~\ref{tab:ablation} compares the performance of different module features. 
\textbf{Risk} is our proposed method of mapping Euler distance to risk space by~(\ref{eq:maprisk}). 
\textbf{LN} is to replace the Probabilistic module with a Linear scalar and bias. \textbf{TV} is to use temporal parameters on the planning time horizon. 
No check mark under this item means the model adopts a constant parameter along each time step of sampled trajectories.
Different sampling count is tested to verify the expendability of our model. 
The sampling count only changes the sampling density of a fixed speed and steering range.
From rows 1, 2, and 4, we can see that our proposed method with the same sampling count performs the best on ADE, FDE~(Lon.), and jerk metrics. 
The time variational mapping parameters can increase the performance of AED by 7.4\%. And the ADE incrementation of the probabilistic model is 84.2\% w.r.t. linear model. 
Rows 3, 4, and 5 show that the performance of the map model will be influenced by downstream planning methods, higher sampling density will achieve better performance without additional training process, but users need to balance the performance and computation resources.

\subsection{Ablation Study of Loss Items}
The results of the Loss items ablation study are listed in Table~\ref{tab:AblationLoss}. Only one item is removed in each ablation study. The baseline model which adopts all the loss items is listed in row 4 of Table~\ref{tab:ablation}. 
Removing \textbf{Individual Velocity Regression} means that the model removes the loss of directly learning velocity value from ground truth as introduced in~(\ref{eq:idr}). 
The results shows that our model convergent in freezing robot status.
From the table we find \textbf{Cross Entropy} loss can improve ADE performance by 22.2\%, \textbf{Cost Consistency} loss item makes the progress of 14.8\%. 
\textbf{Cost Regression}, in theory, will make the results of the model more robust, but in our small data quantity experiments, we obtain better results than the baseline model of 4.6\%. 
Minimizing \textbf{Demo Cost} is an intuitive loss item, but the ADE performance increases by 9.2\% with the absence of it. 
This target may lead to a local minimum in the optimization process of the model.

\section{Conclusion}


In this work, we found an efficient multi-modal prediction result utilization method for local planning by unifying it together with other driving contexts and diagnosing it as RiskMap.
An interpretable and hands-free parameter tuning planner is adopted to demonstrate the utilization and efficiency of RiskMap.
We propose the training and planning framework and analyze the efficiency of model modules and training tools.
This model can be further used in driver behavior analysis and customized agent behavior in simulators.
However, the effectiveness of the use of cost-map-based methods needs to be validated in future works.
The learning-based planners are also supposed to have better generalizability as RiskMap offers extracted expressions of map information.

\bibliographystyle{IEEEtran}
\bibliography{IEEEabrv,ref}

\end{document}